%% file: main.tex
\documentclass[conference]{IEEEtran}
\input{macros}

\begin{document}

\title{Learning Optimal Defender Strategies for CAGE-2 using a POMDP Model
}

\author{\IEEEauthorblockN{Duc Huy Le, Rolf Stadler}
\IEEEauthorblockA{KTH Royal Institute of Technology, Stockholm, Sweden \\
Email: \{dhle, stadler\}@kth.se} \\

}

\maketitle

\begin{abstract}
CAGE-2 is an accepted benchmark for learning and evaluating defender strategies against cyberattacks. It reflects a scenario where a defender agent protects an IT infrastructure against various attacks. Many defender methods for CAGE-2 have been proposed in the literature. In this paper, we construct a formal model for CAGE-2 using the framework of Partially Observable Markov Decision Process (POMDP). Based on this model, we define an optimal defender strategy for CAGE-2 and introduce a method to efficiently learn this strategy. Our method, called BF-PPO, is based on PPO, and it uses particle filter to mitigate the computational complexity due to the large state space of the CAGE-2 model. We evaluate our method in the CAGE-2 CybORG  environment and compare its performance with that of CARDIFF, the highest ranked method on the CAGE-2 leaderboard. We find that our method outperforms CARDIFF regarding the learned defender strategy and the required training time. 

\end{abstract}



\section{Introduction}
\input{parts/introduction}


\section{Related Work}
\label{sec:related-work}

\input{parts/related_work}


\section{Intrusion Response Use Case: CAGE-2}
\label{sec:cage-2}
\input{parts/cage-2}


\section{Formalising CAGE-2 with a Partially Observable Markov Decision Process model}
\label{sec:formal-pomdp}
\input{parts/pomdp}


\section{Evaluating BF-PPO in CAGE-2}
\label{sec:evaluation}
\input{parts/evaluation}


\section{Conclusion and Future Work}
\label{sec:conclusion}

\input{parts/conclusion}


\section{Acknowledgement}
This work has been supported by the DARPA CASTLE program through project ORLANDO and by the WASP NEST program through project AIRR. The authors thank KTH researchers Kim Hammar and Xiaoxuan Wang for their constructive comments.


\input{parts/appendix}


\printbibliography


\end{document}

%% file: macros.tex
\usepackage[style=ieee, backend=biber]{biblatex}
\usepackage[T1]{fontenc}

\usepackage{booktabs} 
\usepackage{multirow, multicol}

\usepackage{empheq}
\usepackage{amsmath,amssymb,amsfonts, amsthm, mathtools} 
\usepackage{dsfont}
\usepackage{bbm, mathrsfs} 
\usepackage{algorithm, algorithmic} 
\usepackage{graphicx} 
\usepackage{textcomp}

\usepackage{caption}
\makeatletter
\def\blfootnote{\gdef\@thefnmark{}\@footnotetext}
\makeatother

\usepackage{subcaption}
\usepackage{eqparbox}

\newcommand{\secref}[1]{\S\ref{#1}}

\newtheoremstyle{custom-theorem-style}  
        {3pt}                                               
        {3pt}                                               
        {\normalfont}                               
        {}                                                  
        {\bfseries\itshape}                 
        {\normalfont\bfseries.}         
        {.5em}                                          
        {}                                                  
\theoremstyle{custom-theorem-style}

\newtheorem{problem}{Problem}

\newcommand{\pomdp}[0]{\mathcal{P}}

\newcommand{\textstate}[1]{\texttt{#1}}
\newcommand{\textset}[1]{\mathbf{#1}}
\newcommand{\textaction}[1]{\mathsf{#1}}

\newcommand{\prob}[1]{\mathbbm{P}[#1]}

\DeclareMathOperator*{\argmax}{arg\,max}

%% file: parts/introduction.tex
\blfootnote{DISTRIBUTION STATEMENT A: Approved for public release; dissemination unlimited. This research is supported by the Defense Advanced Research Project Agency (DARPA) through the CASTLE program under Contract No.W912CG23C0029. The views, opinions, and/or findings expressed are those of the authors and should not be interpreted as representing the official views or policies of the Department of Defense or the U.S. Government.}

Traditionally, systems for intrusion detection and response have relied on rule sets that trigger alarms (e.g., \cite{snortids, broids}). The rule sets have been defined and maintained by human experts. The increasing complexity and rapid changes of digital services and infrastructures have made the maintenance of these rule sets a challenging and a time-consuming task. As a response, research efforts into \textit{automated cyberdefence} have started based on the idea that defender strategies can be dynamically learned and then executed by defender agents with minimal human intervention. One can say that the rules are no longer defined by humans, but automatically constructed from observing systems under attack. 

Over the last decade, various learning approaches have been proposed for automated cyberdefence, including those based on reinforcement learning \cite{related-work-cage-3, rl2, rl3}, stochastic modelling \cite{related-work-intrusion-response-1, related-work-intrusion-response-2, related-work-intrusion-response-3}, game theory \cite{gametheory1, gametheory2, gametheory3}, and most, recently, causal inference \cite{kim-cage2, causal1, causal2}.

\begin{figure}[h!]
    \centering
    \includegraphics[width=0.95\linewidth]{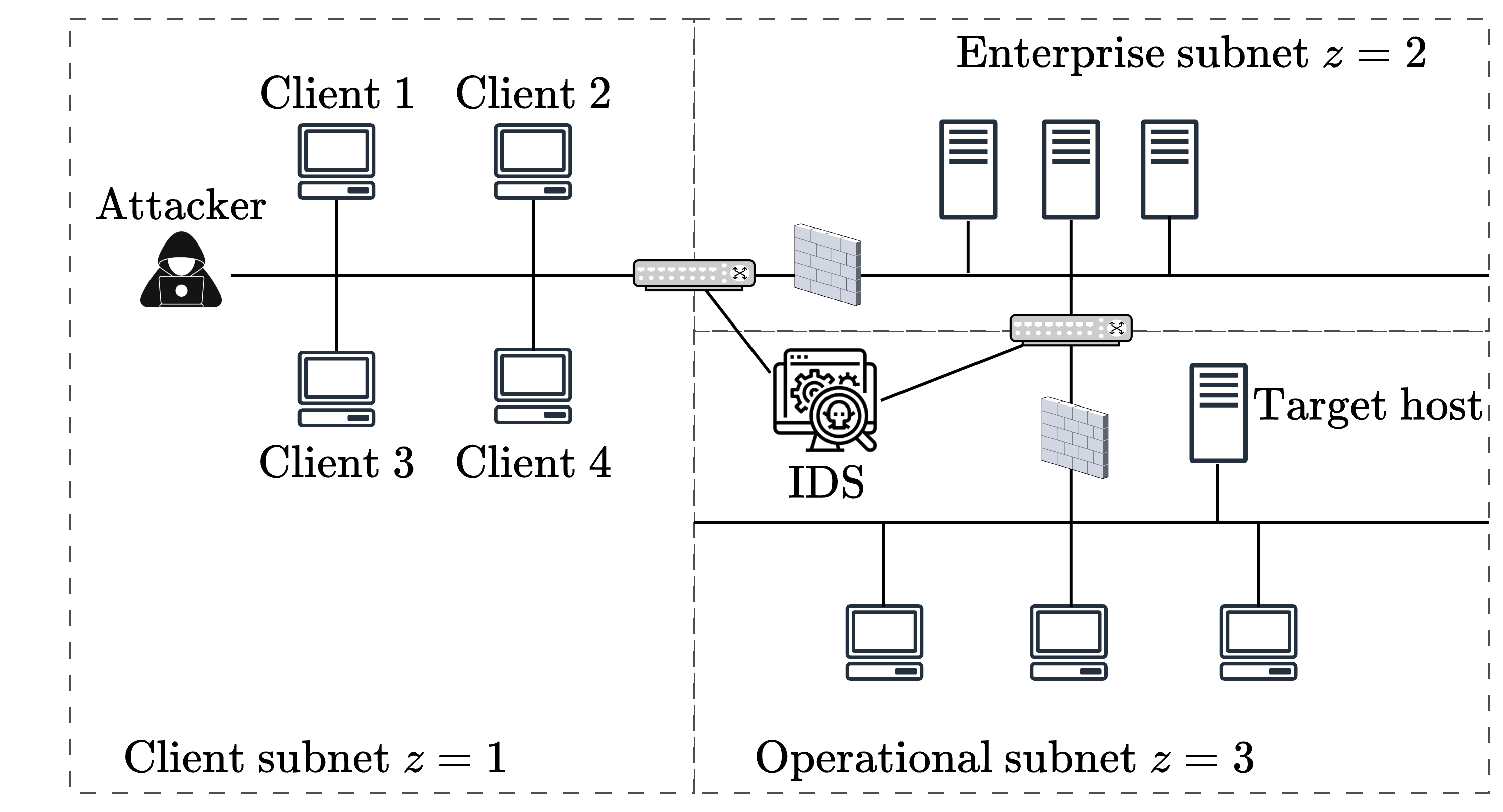}
    \caption{The network topology of CAGE-2 scenario}
    \label{fig:topo}
\end{figure}

Currently, the most popular benchmark environment for learning and evaluating defender strategies is the Cyber Autonomy Gym for Experimentation Challenge 2 (CAGE-2) \cite{cage2}. It is based on a scenario where a defender agent protects an IT infrastructure  against different types of attacks (Fig. \ref{fig:topo}). CAGE-2 includes a simulation environment, named CybORG \cite{cyborg}, in which defender agents can be trained and evaluated.

A large number of defender methods for CAGE-2 have been proposed and published. A CAGE-2 leaderboard ranks the solutions according to a score that measures the effectiveness of the defenders against the attacks. The solutions on the leaderboard are based on heuristic approaches, and none of them is built on a formal model. As a consequence, it is not clear to which extent these solutions are optimal or close to optimal. (There is a recent work that uses a formal model, which we discuss in \secref{sec:related-work}). 


In this paper, we formalise the CAGE-2 scenario and develop a formal model using the framework of Partially Observable Markov Decision Process (POMDP) \cite{pomdp}. The model is obtained from studying the documentation and the source code of CAGE-2, as well as from conducting experiments in the CybORG environment.

The model allows us to define an optimal defender strategy for CAGE-2. To learn this strategy, we propose a learning method, which we call \textit{Belief Filter Policy Proximal Optimisation} (BF-PPO). It is based on the state-of-the-art reinforcement learning algorithm Policy Proximal Optimisation (PPO) \cite{ppo}, and it uses the concept of particle filter \cite{particle_filter} to address the computational complexity arising from the large state space of the CAGE-2 model. We evaluate our method and compare its performance with CARDIFF \cite{cardiff}, the method with the highest rank on the CAGE-2 leaderboard. We find that our method outperforms CARDIFF in terms of the learned defender strategy. Also, our method converges faster, requiring fewer training episodes.

The contributions of this paper are:

\begin{itemize}
    \item We present a formal model of the CAGE-2 scenario using the POMDP framework (presented in \secref{subsec:pomdp-formalisation}) and formally define an optimal defender strategy for CAGE-2 (presented in \secref{subsec:pomdp-defender-strategy}).
    \item We present BF-PPO, a solution method that iteratively approximates the optimal strategy based on PPO and particle filter (presented in \secref{subsec:bf-ppo})
    \item We evaluate BF-PPO in the CybORG environment and show that it outperforms the state-of-the art method CARDIFF, the top performer on the CAGE-2 leaderboard (presented in \secref{sec:evaluation}).
\end{itemize}

%% file: parts/related_work.tex
The CAGE-2 challenge has attracted much attention in the cybersecurity community. More than 35 solution methods have been developed to learn defender strategies for the CAGE-2 scenario, for example \cite{cage2, related-work-cage-1, related-work-cage-2, related-work-cage-3, related-work-cage-4, related-work-cage-7, kim-cage2, related-work-cage-jakob}. All of these methods, except for \cite{kim-cage2}, rely on heuristic techniques and are not based on a formal model. As a consequence, it cannot be shown that any of these approaches lead to an optimal defender strategy. In contrast, we present in this paper a formal model of the CAGE-2 scenario, based on which we define an optimal strategy for the defender, and develop a solution to approximate this strategy.

The only published method that is based on a formal model of the CAGE-2 scenario is included in \cite{kim-cage2}.  The study presents a structural causal model and proposes a defender strategy using Monte Carlo Planning. The method, however, is an online solution and is therefore of a different type than the above cited works, which all use offline training. This paper also uses offline training, which allows us to directly compare its performance with those works.

The POMDP framework has been recently used by many researchers to study intrusion detection and response \cite{related-work-ids-1, related-work-intrusion-response-1, related-work-intrusion-response-2, related-work-intrusion-response-3, related-work-intrusion-response-4, related-work-intrusion-response-5}. These works consider scenarios that are much simpler than CAGE-2 and the corresponding state spaces are much smaller. In order to address the computational complexity that results from larger state spaces, this paper uses the concept of particle filter for state estimation.

%% file: parts/cage-2.tex
 The CAGE-2 scenario presents a cybersecurity use case in which a defender defends an IT infrastructure against an attacker with a fixed strategy. The network contains 11 hosts, divided into 3 subnets (see Fig. \ref{fig:topo}).

The attacker aims to access the target host located in the \textit{operational subnet} and to disrupt the services it provides. After compromising one of the clients in the \textit{client subnet}, the attacker tries to break into one of the hosts in the \textit{enterprise subnet}, before finally attacking the target host. To achieve this goal, it performs a sequence of attack actions of the following types: (1) Discover the hosts on a subnet; (2) Scan for vulnerabilities in a host; (3) Exploit a discovered vulnerability to get access to a host; (4) Escalate the access to become a root; (5) Interrupt the services on the host.

The defender cannot directly observe the progress of the attacker. Instead, it relies on real-time events produced by an Intrusion Detection System (IDS), and responds by taking actions of the following types: (A) Analyse a host to check whether the attacker has performed an action (3), (4) or (5) on it; (B) Deploy a decoy service to deceive the attacker; (C) Neutralise and remove malware installed by the attacker; and (D) Restore the host into a safe state and restart the services. The goal of the defender is balancing the two objectives: maximising the service availability while minimising the access of the attacker.

An attack evolves over a finite number of time steps. During each time, both the attacker and the defender take an action.

%% file: parts/pomdp.tex
We formalise the CAGE-2 scenario introduced in Section \ref{sec:cage-2} using the framework of Partially Observable Markov Decision Process (POMDP) and describe a learning solution method, BF-PPO, which is based on the state-of-the-art reinforcement learning algorithm PPO and the concept of particle filter.

In the following, we use the term CAGE-2 to refer to the CAGE-2 scenario as well as to the implementation of CAGE-2 in the CybORG platform. 


\subsection{Partially Observable Markov Decision Process}

A Partially Observable Markov Decision Process (POMDP) models the progression of a discrete-time stochastic system with partial observability. It is defined by a 9-tuple $\pomdp = ( \textset{S}, \textset{D}, \mathcal{P}^{D_t}_{S_t,S_{t+1}}, \textset{O}, \mathcal{Z}^{D_t,S_{t+1}}_{O_{t+1}}, \mathcal{R}^{D_t}_{S_t}, \gamma, T, \textset{B} )$. $\textset{S}$ denotes the state space and $\textset{D}$ denotes the action space. An episode in a POMDP begins in an initial state $S_t$. At every time $t = 1,\dots,T$ the system performs an action $D_t$, which moves the system state $S_t$ to a new state $S_{t+1}$ with transition probability $\mathcal{P}^{D_t}_{S_t,S_{t+1}} = \mathds{P}[S_{t+1}|S_t, D_t]$. The state transition is partially observable through variable $O_{t} \in \textset{O}$, where $\textset{O}$ is the observation space. The observation function is defined as $\mathcal{Z}^{D_t, S_{t+1}}_{O_{t+1}} = \prob{O_{t+1}|S_{t+1},D_t}$. Associating with a state transition is a reward $R_t = \mathcal{R}^{D_t}_{S_t} \in \mathds{R}$. The objective with a POMDP is to identify a sequence of $T$ actions that maximises the expected cumulative reward $\mathds{E}[J]$, with discount factor $\gamma \in (0,1]$:

\begin{equation}
    J = \sum_{t=1}^T \gamma^{t-1} R_t
\end{equation}

A belief state $b_t = \langle b_t(S_t) \rangle_{S_t \in \textset{S}}$ is associated with time $t$, where $b_t(s) = \prob{S_t = s | h_t}$ with $h_t = (S_1,D_1, O_2, \dots, D_{t-1}, O_t)$. The belief state is a distribution over the state space $\textset{S}$. At every time $t$, the belief is recursively computed:

\begin{equation}
\label{eq:belief-update}
b_{t}(S_{t}) = \eta \mathcal{Z}^{D_{t-1}, S_{t}}_{O_{t}} \sum_{S_{t-1} \in \textset{S}} \mathcal{P}^{D_{t-1}}_{S_{t-1},S_t} b_{t-1}(S_{t-1}) 
\end{equation}

\noindent where $\eta = \sum_{S_t \in \textset{S}} {\mathcal{Z}^{D_{t-1}, S_{t}}_{O_{t}} \sum_{S_{t-1} \in \textset{S}}\mathcal{P}^{D_{t-1}}_{S_{t-1},S_t} b_{t-1}(S_{t-1}) }$ is the normalisation factor. The initial belief state is $b_1(S_1) = 1$ (the initial state is observable).


\subsection{Formalising CAGE-2 using POMDP}
\label{subsec:pomdp-formalisation}

We formulate an attack in the CAGE-2 scenario as a POMDP episode. The POMDP explicitly models the state of the infrastructure, the defender action and the observation produced by the IDS. At time $t$, both the defender and the attacker each perform an action.

\subsubsection{Infrastructure model} 
\label{subsec:pomdp-infrastructure}

Let $\textset{H}$ be the set of $n$ hosts, $\textset{Z}$ be the set of subnets, and $\textset{E}$ be the set of $m$ services. Each host $h \in \textset{H}$ belongs to a subnet defined by $z(h)$. $\textset{E_h} \subset \textset{E}$ denotes the set of services provided by host $h$. Exploiting a service $e \in \textset{E_h}$ grants the attacker one of the following accesses to the host: $\textstate{N}$ (No access), $\textstate{U}$ (User access), and $\textstate{S}$ (Superuser access). The access is determined by function $t(h,e)$. In the CAGE-2 scenario, a host $h_1$ where the attacker has root access provides the attacker with the knowledge of a different host $h_2$. We model this fact with a function $g_M$ that maps $h_1$ to $h_2$. For details of the infrastructure model, see Appendix.

\subsubsection{State space $\textset{S}$} \label{subsec:pomdp-system-state}

The system state $S_t$ represents the collective states of all hosts at time $t$. Formally, $S_t = ( S_{1,t}, \dots, S_{n,t})$, where $S_{h,t}$ is the local state of host $h$. $S_{h,t}$ has three components: the attacker access state $I_{h,t}$, the running services $E_{h,t}$, and the scanned services $F_{h,t}$.

$I_{h,t}$ represents the attacker access to host $h$ and takes one of the following vales: \textstate{H} if the host is unknown to the attacker; \textstate{K} if the host is known to the attacker; \textstate{S} if the host has been scanned by the attacker; \textstate{U} or \textstate{R} if the attacker has performed a successful exploit on the host; \textstate{P} if the attacker has root access on the host; and \textstate{I} if the services on the host are interrupted by the attacker. 

$E_{h,t}$ represents the set of services deployed on host $h$ and $F_{h,t}$ represents the knowledge of the attacker about the services running on host $h$. $E_{h,t}$ and $F_{h,t}$ are subsets of $\textset{E}$. The state space $\textset{S}$ can be written as $\{\{\textstate{H}, \textstate{K}, \textstate{S}, \textstate{U}, \textstate{R}, \textstate{P}, \textstate{I}\} \times 2^{\textset{E}} \times 2^{\textset{E}}\} ^n$.

An episode starts with $S_{h,1} = (I_{h,1} = \textstate{H}, E_{h,1} = \textset{E_h}, F_{h,1} = \emptyset)$ for all hosts $h \in \textset{H}$.


\subsubsection{Action space $\textset{D}$}
\label{subsec:pomdp-defender-action}

At time $t$, the defender takes the action $D_t = (\Delta_t, T_t)$, whereby $\Delta_t$ is the action type, namely, \textit{Analyse} a host ($\textaction{A}$); Deploy a decoy service $e \in \textset{E}$ ($\textaction{D}_e$); \textit{Neutralise} and remove malware on a host ($\textaction{N}$); and \textit{Restore} the host to a secure state and restart the services ($\textaction{R}$). $T_t$ is the target host. In addition, the defender has the option to perform no action at time $t$, in which case we write $D_t = \textaction{I}$. Thus, the action space is $\textset{D} = \textaction{I} \cup \{ \textaction{A}, \textaction{D}_1,\dots,\textaction{D_m}, \textaction{N}, \textaction{R}\} \times \textset{H}$, where $D_1,\dots, D_m$ are the decoy actions for each of the $m$ services in $\textset{E}$.


\bgroup
\begin{figure*}[htbp]
    \centering
    \includegraphics[width=0.85\linewidth]{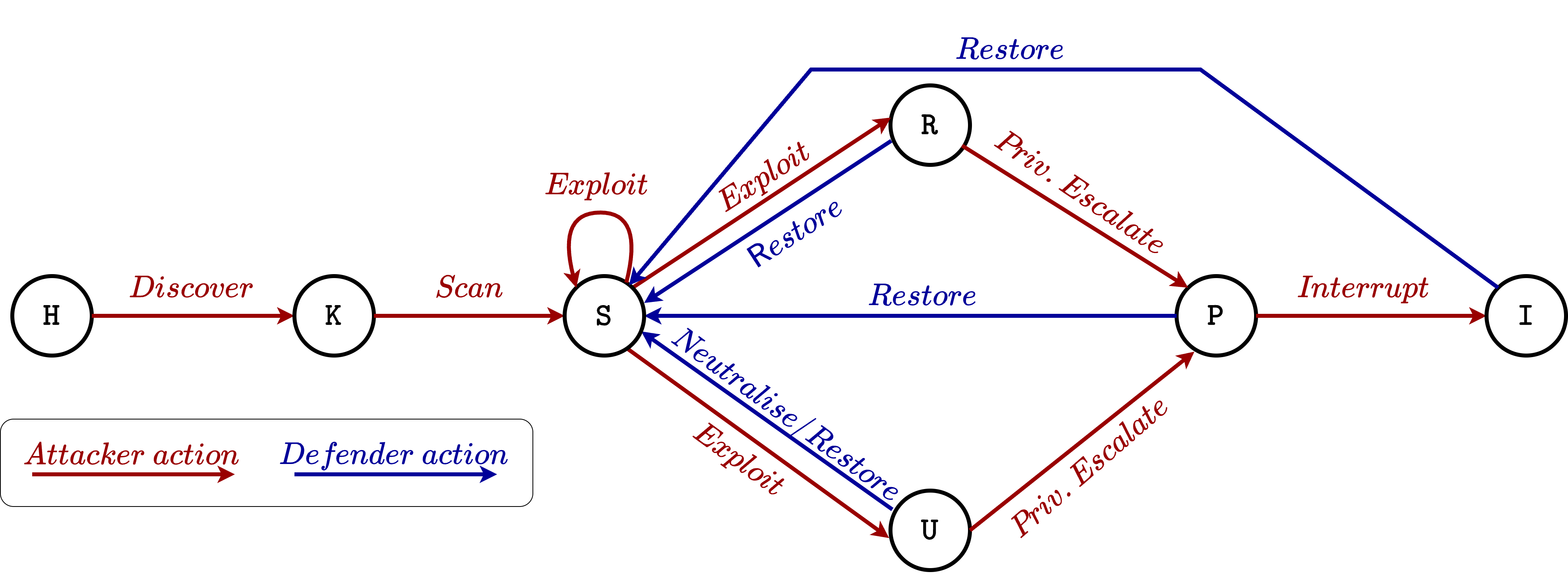}
    \caption{The transition of the attacker access state $I_{h,t}$ caused by an action from the attacker $A_t$ or the defender $D_t$; nodes present the access states; arrows present the actions that cause state transitions. The defender actions \textit{Analyse} and \textit{Decoy} do not change the state $I_{h,t}$ and they are therefore not included.}
    \label{fig:state_transition}
\end{figure*}
\egroup

\subsubsection{State transition}
\label{subsec:formalisation-state-trans}

The attacker in CAGE-2 follows a fixed strategy $\pi_A$ that maps the state $S_t$ to an action $A_t$ at time $t$. We model $A_t = (\Lambda_t, T_t)$, where $\Lambda_t$ is an action type, namely,  \textit{Discover} ($\textaction{D}$); \textit{Scan} ($\textaction{S}$); \textit{Exploit} service $e \in \textset{E}$ ($\textaction{E}_e$); \textit{Priviledge escalate} ($\textaction{P}$); and \textit{Interrupt} ($\textaction{I}$) (see \secref{sec:cage-2}). $T_t$ is the target of $\Lambda_t$, which is either a subnet $z \in \textset{Z}$ (for action $\textaction{D}$) or a host $h \in \textset{H}$ (for other actions). 

In CAGE-2, we can describe the state transition $S_t \rightarrow S_{t+1}$ through the transition of the host states $S_{h,t} \rightarrow S_{h,t+1}, \; h \in \textset{H}$. Fig. \ref{fig:state_transition} shows the state transition diagram of a host. 

At time $t=1,\dots,T$, the defender performs an action $D_t$, followed by the attacker performing an action $A_t$. The sequencing may look strange, but this is the way CAGE-2 is designed. We can therefore decompose the transition $S_{h,t} \rightarrow S_{h,t+1}$ into two consecutive steps. 

\paragraph{$S_{h,t}\rightarrow S'_{h,t}$ (defender action)}

\begin{subequations}
\label{eq:pomp-trans:defender}
\begin{align}
\label{eq:pomdp-trans:defender:neutralise}I_{h,t} = \textstate{U} &\rightarrow I'_{h,t} = \textstate{S}  \text{ if } D_t = (\textaction{N},h) \\
\label{eq:pomdp-trans:defender:restore} I_{h,t} \in \{ \textstate{U}, \textstate{R}, \textstate{P}, \textstate{I} \} &\rightarrow I'_{h,t} = \textstate{S} \text{ if } D_t = (\textaction{R},h) \\
\label{eq:pomdp-trans:defender:otherwise} I_{h,t} &\rightarrow I'_{h,t} = I_{h,t} \text{ otherwise}
\end{align}
\end{subequations}

(\ref{eq:pomdp-trans:defender:neutralise} - \ref{eq:pomdp-trans:defender:otherwise}) describe the transition of the attacker access state $I_{h,t} \rightarrow I'_{h,t}$. (\ref{eq:pomdp-trans:defender:neutralise}) captures the effect of the \textit{Neutralise} action, which removes the attacker from the host if $I_{h,t} = \textstate{U}$. (\ref{eq:pomdp-trans:defender:restore}) states that the \textit{Restore} action sets the host in the secure state $\textstate{S}$ (see Fig. \ref{fig:state_transition}).

\begin{subequations}
\label{eq:pomdp-trans:defender:decoy}
\begin{align}
\label{eq:pomdp-trans:defender:decoy-success} E_{h,t} &\rightarrow E'_{h,t} = E_{h,t} \cup \{e\} \text{ if } D_t = (\textaction{E}_e,h), e \notin E_{h,t} \\
\label{eq:pomdp-trans:defender:decoy-restore} E_{h,t} &\rightarrow E'_{h,t} = \textset{E_h} \text{ if } D_t = (\textaction{R},h) \\
\label{eq:pomdp-trans:defender:decoy-otherwise} E_{h,t} &\rightarrow E'_{h,t} = E_{h,t} \text{ otherwise}
\end{align}
\end{subequations}

(\ref{eq:pomdp-trans:defender:decoy-success}-\ref{eq:pomdp-trans:defender:decoy-otherwise}) describe the transition of the running services $E_{h,t} \rightarrow E'_{h,t}$. (\ref{eq:pomdp-trans:defender:decoy-success}) presents the installation of a new decoy service on host $h$. (\ref{eq:pomdp-trans:defender:decoy-restore}) presents the removal of all decoy services as an effect of the \textit{Restore} action. 

Lastly, the scanned services component is not affected by the defender action, i.e., $F'_{h,t} = F_{h,t} \; \forall D_t \in \textset{D}$.

\paragraph{$S'_{h,t} \rightarrow S_{h,t+1}$ (attacker action)}

\begin{subequations}
\label{eq:pomp-trans:attacker}
\begin{align}
\nonumber &I'_{h,t} = \textstate{H} \rightarrow  I_{h,t+1} = \textstate{K} \text{ if }\\
\label{eq:pomdp-trans:attacker:from_h_to_k}
&\begin{cases}
A_t = (\textaction{D}, z(h) = 1) \\
A_t = (\textaction{D}, z(h)) \: \exists h': z(h') = z(h), I'_{h',t} \in \{\textstate{P},\textstate{I} \} \\
\exists h': I'_{h',t} = \textstate{P}, g_M(h') = h 
\end{cases} \\
\label{eq:pomdp-trans:attacker:from_k}
&I'_{h,t} =  \textstate{K} \rightarrow  I_{h,t+1} = \textstate{S} \text{ if } A_t = (\textaction{S},h)  \\
\nonumber &I'_{h,t} =  \textstate{S} \rightarrow I_{h,t+1} = \\
\label{eq:pomdp-trans:attacker:from_s}
&\begin{cases}
    \textstate{S} \text{ if } A_t=(\textaction{E}_e,h), e \notin \textset{E_h} \textit{ or } t(h,e) = \textstate{N}  \\
    \textstate{U} \text{ if } A_t=(\textaction{E_e},h), e \in \textset{E_h}, t(h,e) = \textstate{U} \\
    \textstate{R} \text{ if } A_t=(\textaction{E_e},h), e \in \textset{E_h}, t(h,e) = \textstate{S} 
\end{cases} \\
\label{eq:pomdp-trans:attacker:to_p_1}
&I'_{h,t} = \textstate{U} \rightarrow I_{h,t+1} = \textstate{P} \text{ if } A_t=(\textaction{P},h) \\
\label{eq:pomdp-trans:attacker:to_p_2}
&I'_{h,t} = \textstate{R} \rightarrow I_{h,t+1} = \textstate{P} \text{ if } A_t=(\textaction{P},h) \\
\label{eq:pomdp-trans:attacker:to_i}
&I'_{h,t} =  \textstate{P} \rightarrow  I_{h,t+1} = \textstate{I} \text{ if } A_t=(\textaction{I},h) \\
\label{eq:pomdp-trans:attacker:otherwise} 
&\; \; \; \; \; \;\; I'_{h,t} \rightarrow I_{h,t+1} = I'_{h,t} \text{ otherwise}
\end{align}
\end{subequations}

(\ref{eq:pomdp-trans:attacker:from_h_to_k}-\ref{eq:pomdp-trans:attacker:otherwise}) describe the transition of the attacker access state $I'_{h,t} \rightarrow I_{h, t+1}$. (\ref{eq:pomdp-trans:attacker:from_h_to_k}) describes the three cases causing the transition $\textstate{H} \rightarrow \textstate{K}$, where host $h$ becomes exposed to the attacker. The first case presents the initial phase of an attack with the \textit{Discover} action on the clients on subnet $z=1$. The second case represents the \textit{Discover} action other subnets $z=2,3$, the attacker must have root access to another host on that subnet. The attacker can also discover host $h$ by having root access to host $h'$ that is connected to $h$, i.e., $g_M(h') = h$ (the last case). (\ref{eq:pomdp-trans:attacker:from_k}) describes the transition $\textstate{K} \rightarrow \textstate{S}$, where the attacker scans a known host $h$ for vulnerabilities. 

After scanning host $h$ ($I'_{h,t} = \textstate{S}$), the attacker attempts to gain access by exploiting one of the scanned services $e \in F'_{h,t}$. (\ref{eq:pomdp-trans:attacker:from_s}) describes the three possible outcomes of this action, which are illustrated in Fig. \ref{fig:state_transition}. The exploit fails if the target service is a decoy ($e \notin \textset{E_h}$) or not exploitable ($t(h,e)$ = \textstate{N}) (first case of (\ref{eq:pomdp-trans:attacker:from_s})). Otherwise, a successful exploit, presented by the last two cases of (\ref{eq:pomdp-trans:attacker:from_s}), gives the attacker access to host $h$ depending on the access right of the target service, either a regular user ($t(h,e) = \textstate{U}$) or a SuperUser ($t(h,e) = \textstate{S})$). Finally, (\ref{eq:pomdp-trans:attacker:to_p_1}) and (\ref{eq:pomdp-trans:attacker:to_p_2}) describe the attacker gaining root access, and (\ref{eq:pomdp-trans:attacker:to_i}) describes the attacker interrupting the services running on host $h$.

\begin{subequations}
\label{eq:pomdp-trans:attacker:scan}
\begin{align}
\label{eq:pomdp-trans:attacker:scan-success} F'_{h,t} &\rightarrow F_{h,t+1} = E'_{h,t} \text{ if } A_t = (\textaction{S},h) \\
\label{eq:pomdp-trans:attacker:scan-otherwise} F'_{h,t} &\rightarrow F_{h,t+1} = F'_{h,t} \text{ otherwise}
\end{align}
\end{subequations}

(\ref{eq:pomdp-trans:attacker:scan-success}) and (\ref{eq:pomdp-trans:attacker:scan-otherwise}) describe the transition of the scanned services $F'_{h,t} \rightarrow F_{h,t+1}$, which is only triggered when the attacker perform the \textit{Scan} action to explore the services running on host $h$ (\ref{eq:pomdp-trans:attacker:scan-success}). 

Lastly, the running services component is not affected by the attacker action, i.e., $E_{h,t+1} = E'_{h,t} \; \forall A_t$.

\subsubsection{Time horizon $T$} 

An episode in CAGE-2 has a finite time horizon $T$.


\subsubsection{Observation space $\textset{O}$} 
\label{subsec:pomdp-observations}

At time t, the defender observes $O_t = ( O_{1,t}, \dots, O_{n,t})$. The host observation $O_{h,t}$ takes one of the following values: $\textstate{H}$ if the host has not been scanned by the attacker; $\textstate{S}$ if the host has been scanned at time $t-1$ by the attacker; $\textstate{C}$ if the IDS has issued an alarm for the host; $\textstate{P}$ if the attacker has root access to the host; $\textstate{U}$ if the $\textaction{N}$\textit{eutralise} action has been performed by the defender; and $\textstate{N}$ if there is no detected attacker activity on a scanned host. Thus, the observation space is $\textset{O} = \{\textstate{H}, \textstate{S}, \textstate{C}, \textstate{P}, \textstate{U}, \textstate{N}\}^n$. The initial observation $O_{h,1} = \textstate{H}, \; \forall \, h \in \textset{H}$.

\subsubsection{Observation function}
\label{subsec:pomdp-observation-function}

We describe the observation function for $O_t$ at time $t=2,\dots,T$ using the host observations $O_{h,t}$, which depend on the host state $S_{h,t}$, the attacker action $A_{t-1}$ and the defender action $D_{t-1}$:

\begin{subequations}
\label{eq:pomdp-observation-function}
\begin{empheq}[left={O_{h,t} = \empheqlbrace \;}]{alignat=2}
\label{eq:pomdp-observation-function-h} 
\textstate{H} &\text{ if } I_{h,t} \in \{\textstate{H},\textstate{K} \} \\
\label{eq:pomdp-observation-function-s-1} 
\textstate{S} &\text{ if } A_{t-1} = (\textaction{S}, h) \\
\label{eq:pomdp-observation-function-s-2} 
\textstate{S} &\text{ if } A_{t-1} = (\textaction{E_e},h), N_{d} = 0 \; \forall e \in \textset{E} \\
\label{eq:pomdp-observation-function-c-1} 
\textstate{C} &\text{ if } A_{t-1} = (\textaction{E_e},h), N_{d} = 1 \; \forall e \in \textset{E}\\
\label{eq:pomdp-observation-function-u} 
\textstate{U} &\text{ if } D_{t-1} = (\textaction{N},h)\\
\label{eq:pomdp-observation-function-n-1} 
\textstate{N} &\text{ if } D_{t-1} = (\textaction{R},h)\\
\label{eq:pomdp-observation-function-c-2} 
\textstate{C} &\text{ if } I_{h,t} \in \{\textstate{U},\textstate{R}\}, D_{t-1} = (\textaction{A},h)\\
\label{eq:pomdp-observation-function-p} 
\textstate{P} &\text{ if } I_{h,t} \in \{\textstate{P},\textstate{I}\}, D_{t-1} = (\textaction{A},h)\\
\label{eq:pomdp-observation-function-n-2} 
\textstate{N} &\text{ otherwise }
\end{empheq}
\end{subequations}

\noindent where $N_d$ is a binary random variable that determines the observation $O_{h,t}$ when the attacker performs the \textit{Exploit} action on host $h$. 

In (\ref{eq:pomdp-observation-function-h}) and (\ref{eq:pomdp-observation-function-s-1}), $O_{h,t}$ refers to the observations before and after the attacker scans host $h$, respectively. After the attacker performs the \textit{Exploit} action at time $t-1$, $O_{h,t}$ has value $\textstate{S}$ if the exploitation is not detected by the IDS (\ref{eq:pomdp-observation-function-s-2}), otherwise, it has value $\textstate{C}$ (\ref{eq:pomdp-observation-function-c-1}). (\ref{eq:pomdp-observation-function-u}) and (\ref{eq:pomdp-observation-function-n-1}) describe the observations produced by the defender actions \textit{Neutralise} and \textit{Restore}, respectively. The defender can perform the \textit{Analyse} action to learn about the compromise state of host $h$ (\ref{eq:pomdp-observation-function-c-2}-\ref{eq:pomdp-observation-function-p}). Otherwise, the observation $O_{h,t} = \textstate{N}$ (\ref{eq:pomdp-observation-function-n-2}).


\subsubsection{Reward function $\mathcal{R}^{A_t}_{S_t}$} \label{subsec:pomdp-reward}

In CAGE-2, upon performing an action at time $t$, the defender receives a reward $R_t$:

\begin{equation}
\label{eq:pomdp:reward}
    R_t = \sigma_{D_t} + \sum_{h \in \textset{H}} (\sigma_{z(h)} \mathds{1}_{I_{h,t} \in \{\textstate{U}, \textstate{R}, \textstate{P}, \textstate{I}\}}
    +  \sigma_h \mathds{1}_{I_{h,t} =\textstate{I}} )
\end{equation}

\noindent where $\sigma_{\textaction{D_t}} < 0$ defines the reward (actually, the cost) for performing the action $D_t$; $\sigma_{z(h)} < 0$ is the reward for each compromised host in subnet $z(h)$; and $\sigma_h < 0$ is the reward for service interruption on host $h$. The values of these parameters in CAGE-2 are presented in the Appendix.


\subsection{Defender Problem}
\label{subsec:pomdp-defender-strategy}

The objective of the defender in CAGE-2 is to maximise the expected cumulative reward $J$ with the discount factor $\gamma = 1$: 

\begin{equation}
\label{eq:pomdp-obj-func}
    J(\pi_D) = \sum_{t=1}^T \mathds{E}_{\pi_D} [R_t]
\end{equation} 

\noindent whereby $\pi_D$ is the defender strategy, which defines a mapping from the belief space to the action space. 

Therefore, the defender problem is to find the optimal strategy $\pi_D^*$ that maximises the expected cumulative reward over the time horizon $T$.

\begin{problem}
\label{prob:pomdp}
Find the optimal defender strategy under the POMDP model of CAGE-2:
\begin{subequations}
\label{eq:pomdp-opt-problem}
\begin{align}
    \pi_D^* \; \; \;\;\;& = \; \argmax_{\pi_D} \mathds{E}_{\pi_D} [J] \\
    \text{subject to } \; \; \;\;\; & D_t = \pi_D(b_t)  \forall t \\
                       & \pi_A \sim P(\Pi_A)
\end{align}
\end{subequations}
\end{problem}

As the POMDP is stationary with a finite time horizon, we know that an optimal strategy $\pi^*_D$ exists \cite[Thm. 7.4.1]{pomdp}.

\bgroup
\begin{figure*}[htbp]
    \centering
    \includegraphics[width=0.92\linewidth]{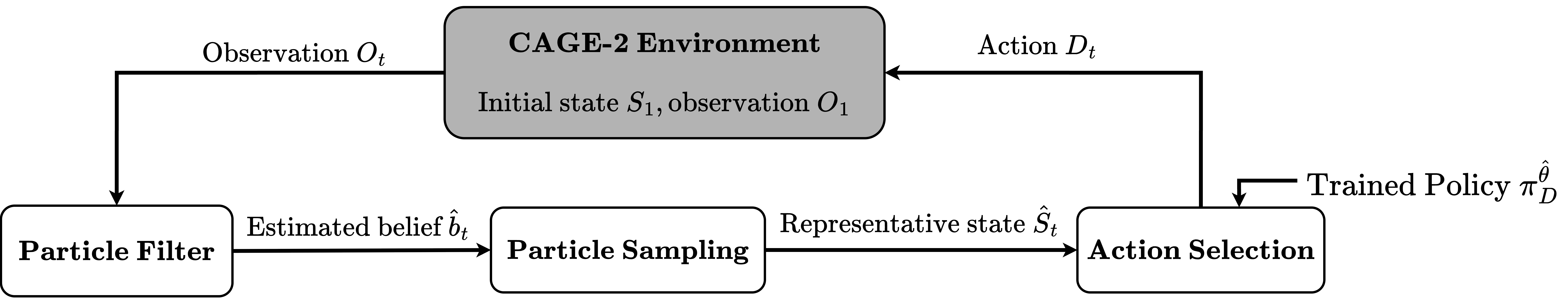}
    \caption{Belief Filter Proximal Policy Optimisation (BF-PPO)}
    \label{fig:solutions}
\end{figure*}
\egroup


\subsection{Computing the optimal defender strategy: BF-PPO}
\label{subsec:bf-ppo}

The optimal strategy $\pi^*_D$ can be computed by \textit{dynamic programming} methods such as value iteration \cite{dynamic-programming-pomdp}. However, the large size of the state space (in the order of $10^{39}$) leads to a high-dimensional belief. As a result, exact computation or estimation of $\pi^*_D$ with the mentioned methods are computationally intractable \cite{pomdp-complexity1, pomdp-complexity2}. 

We therefore apply an iterative approximation strategy using Reinforcement Learning in form of Proximal Policy Optimization (PPO) \cite{ppo}. It uses a neural network to represent the policy and performs policy optimisation using gradient ascent with a clipped objective function to prevent large policy updates.

In this work, we make two additions to the traditional algorithm regarding the estimation of the belief state and its representation.
First, the evolution of a POMDP requires a belief update every time $t$, which is calculated with the Bayes Filter (\ref{eq:belief-update}). The update has a quadratic time complexity with respect to the size of the state space $\textset{S}$. For that reason, the Bayes Filter is computationally impractical in our case. We address the issue by updating the belief state using particle filter \cite{particle_filter}, which is an approximate, non-parametric implementation of the Bayes Filter. The method approximates $b_t$ with a set of $M$ random state samples (or \textit{particles}), denoted as $\mathcal{P}_t = \{s_t^{(1)},\dots,s_t^{(M)}\}$ with $s_t^{(i)} \in \textset{S}$. The belief state (\ref{eq:belief-update}) is estimated using the frequency of the particle states in $\mathcal{P}_t$, i.e., $\hat{b_t}(s_t) =  \frac{1}{M}\sum_{i}^M \mathds{1}_{s_t^{(i)} = s_t}$. The particles at time $t$ are sampled using rejection sampling, as presented in Alg. \ref{alg:particle-filter}.




\begin{algorithm}
\begin{small}
\caption{Particle filter in CAGE-2}\label{alg:particle-filter}
\begin{algorithmic}[1]
\STATE \textbf{Input:} 
  Input particle set $\mathcal{P}_{t-1} = \{s_{t-1}^{(1)}, \dots, s_{t-1}^{(M)}\}$;
  Defender action $D_{t-1}$;
  Observation $O_t$;
  Simulator $\mathscr{S}$;
  $\mathcal{P}_t = \emptyset$ \label{alg:particle-filter:input}

\IF{$t=1$}
    \STATE $\mathcal{P}_{t} \leftarrow \{S_1\}$ \label{alg:particle-filter:init} \\
\ELSE
    \WHILE{$|\mathcal{P}_{t}| < M$}
        \STATE  $\bar{s} \sim \text{Uniform}(\mathcal{P}_{t-1}$)  \label{alg:particle-filter-step1-1}\\ 
        \STATE Set $\mathscr{S}$ to state $\bar{s}$ \\
        \STATE State $\bar{S}_t$, Observation $\bar{O}_t \leftarrow$ execute $D_{t-1}$ on $\mathscr{S}$ \label{alg:particle-filter-step1-n}\\ 
        \IF{$\bar{O}_t = O_t$} \label{alg:particle-filter-step2-1}
           \STATE  $\mathcal{P}_{t} \leftarrow \mathcal{P}_{t} \cup \{\bar{S}_t\}$
        \ENDIF \label{alg:particle-filter-step2-n}
    \ENDWHILE
\ENDIF
\STATE \textbf{Output:} Set of particles $\mathcal{P}_{t}$ 
\end{algorithmic}
\end{small}
\end{algorithm}

Line \ref{alg:particle-filter:init} of Alg. \ref{alg:particle-filter} presents the initial particle set at the beginning of an episode, where the defender has full knowledge of the system state $S_1$. Otherwise, the sampling routine consists of two parts. The first part (lines \ref{alg:particle-filter-step1-1}-\ref{alg:particle-filter-step1-n}) samples the candidate particle states of $\mathcal{P}_t$ by executing action $D_{t-1}$ on each particle state $s_{t-1}^{(i)} \in \mathcal{P}_{t-1}$. The second part (lines \ref{alg:particle-filter-step2-1}-\ref{alg:particle-filter-step2-n}) concentrates the particles in states that are most likely to generate the observation $O_t$. 

Second, based on the output of particle filter, we find a representation of the belief state to be used as input to the neural network encoded strategy. We cannot use $\{b_t(S_t)\}_{S_t \in \textset{S}}$ as the representation since the input layer of the neural network would be very large ($\sim 10^{39}$ in our case). Instead, we decide to represent the belief state through a representative sample state, which is sampled from $\mathcal{P}_t$ (An alternative would have been taking the most likely particle state from $\mathcal{P}_t$):

\begin{equation}
\label{eq:pomdp-particle-aggregation}
   \hat{S}_{t} \sim \text{Uniform}(\mathcal{P}_t)
\end{equation}

PPO, particle filter and particle sampling are the key elements of our solution method, which we call Belief Filter Policy Proximal Optimisation (BF-PPO). At time $t$, the method selects action $D_t$ in three steps, which is illustrated in Fig. \ref{fig:solutions} and detailed in Alg. \ref{alg:bf-ppo}. First, it approximates the belief state $b_t$ with particle filter (Alg. \ref{alg:bf-ppo}, Line \ref{alg:bf-ppo:step1}). From the output of the particle filter $\mathcal{P}_t$, it samples a representative particle state $\hat{S}_t$ (Alg. \ref{alg:bf-ppo}, Line \ref{alg:bf-ppo:step2}). Lastly, $\hat{S}_t$ is input to a neural network that is trained with PPO to generate action $D_t$ (Alg. \ref{alg:bf-ppo}, Line \ref{alg:bf-ppo:step3}). The policy $\pi_D^{\hat{\theta}}$ used in the action selection step is trained through PPO, presented in Alg. \ref{alg:bf-ppo-train}.

\begin{algorithm}[ht]
\begin{small}
\caption{Action Selection with BF-PPO}\label{alg:bf-ppo}
\begin{algorithmic}[1]
\STATE \textbf{Input:} CAGE-2 simulator $\mathscr{S}$; Particle set $\mathcal{P}_{t-1} = \{s_{t-1}^{(1)},\dots,s_{t-1}^{(M)}\}$; Observation $O_t$; Action $D_{t-1}$; defender strategy $\pi^{\hat{\theta}}_D$ (trained by Alg. \ref{alg:bf-ppo-train})

\STATE $\mathcal{P}_t \leftarrow $ \textit{ParticleFilter}($\mathcal{P}_{t-1},O_{t}, D_{t-1}, \mathscr{S}$) \hfill \COMMENT{Alg. \ref{alg:particle-filter}}  \label{alg:bf-ppo:step1} \\
\STATE $\hat{S}_t \sim \text{Uniform}(\mathcal{P}_t)$ \hfill \COMMENT{Eq. (\ref{eq:pomdp-particle-aggregation})}\label{alg:bf-ppo:step2} \\
\STATE $D_t \leftarrow \pi^{\hat{\theta}}_D$ \label{alg:bf-ppo:step3}\\
\STATE \textbf{Output:} Action $D_t$ 
\end{algorithmic}
\end{small}
\end{algorithm}

\begin{algorithm}[ht]
\begin{small}
\caption{Defender strategy training with BF-PPO}\label{alg:bf-ppo-train}
\begin{algorithmic}[1]
\STATE \textbf{Input:} CAGE-2 simulator $\mathscr{S}$, Time horizon $T$, \# iterations $n_I$, \# training episodes $n_E$, Initial strategy $\pi_D^\theta$

\FOR{$\text{iteration } i \leftarrow 1, \dots, n_I$}
    \STATE \COMMENT{Collect trajectories}  \\
    \STATE Initialise trace buffer $\mathcal{B} \leftarrow \emptyset$ \\
    \FOR{episode $e \leftarrow 1, \dots, n_E$}
        \FOR{$t \leftarrow 1, \dots, T$}
            \STATE $\mathcal{P}_t \leftarrow \textit{ParticleFilter}(\mathcal{P}_{t-1}, O_t, \mathscr{S})$ \hfill \COMMENT{Alg. \ref{alg:particle-filter}}
            \STATE $\hat{S}_t \sim \text{Uniform}(\mathcal{P}_t)$
            \STATE $p_{D_t}, D_t \sim \pi_\theta(\cdot \mid \hat{S}_t)$
            \STATE $O_{t+1}, R_t \leftarrow$ Execute $D_t$ in $\mathscr{S}$
            \STATE Store $(\hat{S}_t, D_t, p_{D_t}, R_t, \hat{S}_{t+1})$ in $\mathcal{B}$
        \ENDFOR
    \ENDFOR

    \STATE \COMMENT{Update $\theta$ with PPO \cite[Alg. 1]{ppo}}
    \STATE $\hat{A} \leftarrow $ Monte Carlo advantage
        \STATE Update $\theta$ using clipped surrogate objective:
        \[
            \theta \leftarrow \theta + \alpha \nabla_\theta \mathbb{E}_{t} \left[ 
            \min\left( \rho_t \hat{A}, \text{clip}(\rho_t, 1 - \epsilon, 1 + \epsilon)\hat{A} \right)
            \right]
        \]
\ENDFOR

\STATE \textbf{Output:} Learned defender strategy $\pi^{\hat{\theta}}_D = \pi^{\theta}_D$
\end{algorithmic}
\end{small}
\end{algorithm}

%% file: parts/evaluation.tex
We evaluate our solution method BF-PPO for the CAGE-2 scenario, and compare its performance with that of a state-of-the-art solution. We implement BF-PPO in Python. The hyperparameters are listed in Appendix.

\subsection{Evaluation setup} 
\label{subsec:evaluation-setup}

\subsubsection{Baseline} 
We compare the performance of BF-PPO with CARDIFF \cite{cardiff}, a state-of-the-art method for CAGE-2, which achieves the highest performance among the methods on the CAGE-2 leaderboard \cite{cage-github}. CARDIFF is not based on a formal model. It combines PPO with hierarchical reinforcement learning. CARDIFF is open source at \cite{cardiff}.

\subsubsection{Evaluation metrics}

We use the average cumulative reward $\hat{J}$, which is the average of cumulative rewards across multiple episodes, as the main evaluation metric.

\subsubsection{Attacker scenarios}

The CAGE-2 challenge includes two main attacker scenarios.

\noindent \textbf{B-LINE.} The defender defends the system against the \texttt{B-LINE} attacker \cite{cage2}. The attacker has prior knowledge of the network topology and the intrusion proceeds directly to the target server (see Fig. \ref{fig:topo}).

\noindent \textbf{MEANDER.} The defender defends the system against the \texttt{Meander} attacker \cite{cage2}. The attacker explores the network topology and attempts to gain privilege access to each host it encounters, until it reaches the target server (see Fig. \ref{fig:topo}).

\bgroup
\begin{figure*}[htbp]
    \centering
    \includegraphics[width=0.95\linewidth]{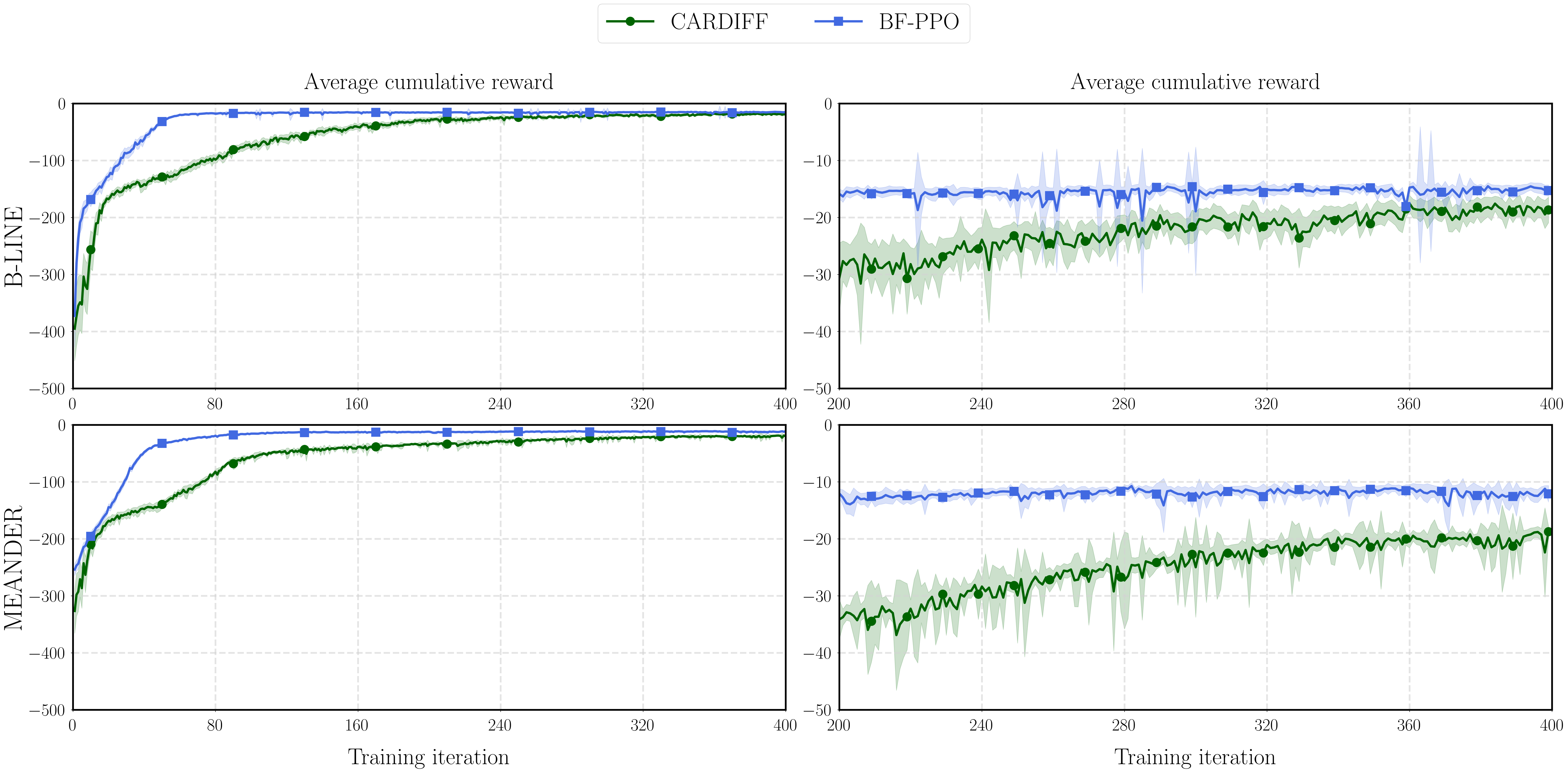}
    \caption{The learning curves for our solution method, BF-PPO (blue curves) and the baseline, CARDIFF (green curves). Each row indicates a CAGE-2 attacker scenario, B-LINE and MEANDER. The left column shows the average cumulative rewards over the training period. The right column enlarges the right half of the graph on the left column. The curves show the mean and the 95\% confidence interval for four training runs with different random seeds.}
    \label{fig:learning-curves}
\end{figure*}
\egroup


\bgroup
\def\arraystretch{1.2} 
\begin{table*}[htbp]
    \centering
    \caption{The evaluation results of our solution method BF-PPO and the baseline CARDIFF for the CAGE-2 scenario. Each subcolumn corresponds to a different time horizon $T$ for an episode. The table cells show the mean and standard deviation of the cumulative rewards, recorded for $100$ episodes.}
    \begin{tabular}{p{0.1\textwidth}|@{\hspace{0pt}}c@{\hspace{1.2pt}}c@{\hspace{1.2pt}}c|@{\hspace{0pt}}c@{\hspace{1.2pt}}c@{\hspace{1.2pt}}c}
    \toprule
    \multirow{2}{*}{ \textbf{Model}} & \multicolumn{3}{c|}{ \textbf{BLINE}} & \multicolumn{3}{c}{ \textbf{MEANDER}} \\
    \cmidrule{2-7}
    &  $T=30$&  $T=50$ &  $T=100$ &  $T=30$ &  $T=50$ &  $T=100$   \\
    \midrule
     CARDIFF & \boldmath{ $-3.41 \pm 1.77$} & 
                { $-6.41 \pm 2.41$} & 
                { $-13.76 \pm 4.25$} & 
                { $-5.64 \pm 1.29$} & 
                { $-8.69 \pm 2.20$} & 
                { $-16.6 \pm 3.83$} \\
    \midrule
     BF-PPO & { $-3.42 \pm 1.35$} & 
            { \boldmath{$-6.24 \pm 2.09$}} & 
            \boldmath{ $-12.82 \pm 3.07$} & 
            \boldmath{ $-4.22 \pm 1.96$} & 
            \boldmath{ $-6.92 \pm 2.84$} & 
            \boldmath{ $-11.56 \pm 4.22$} \\
    \bottomrule
    \end{tabular}
    \label{tab:result} 
\end{table*}
\egroup

\subsubsection{Evaluation Process}

We train BF-PPO and CARDIFF to estimate an optimal defender strategy against each attacker scenario. A training run consists of $400$ iterations. Each iteration corresponds to an update to the strategy parameters and consists of $100$ episodes with time horizon $T=100$. The defender strategies are evaluated at every iteration. We perform four training runs with different random seeds (listed in Appendix). 

Second, we evaluate the defender strategy learned through BF-PPO against each attacker scenario with different time horizons, $T=30,50,100$. Each combination of attacker and time horizon is run for $100$ episodes. We compare the performance of BF-PPO with the published performance of CARDIFF on the CAGE-2 challenge leaderboard \cite{cage-github}.


\subsection{Evaluation results}
\label{subsec:evaluation-results}

Fig. \ref{fig:learning-curves} shows the training performance of the study methods against the two attacker scenarios. The blue curves represent the performance of our solution method BF-PPO. The green curves represent the performance of the baseline CARDIFF. Each row corresponds to the training results against an attacker scenario. The left column shows the cumulative reward of the defender strategies during training run with $400$ iterations. The right column enlarges the second half of the training.

Fig. \ref{fig:learning-curves} shows that the learning curves of BF-PPO quickly converge to a stable mean value for both attackers, indicating that its learned strategies have converged. On the other hand, we observe an increasing trend for the learning curves of CARDIFF throughout the training period, indicating that its strategies have not yet converged. We conclude that our method BF-PPO exhibits significantly faster convergence than CARDIFF.

We also observe that the learning curves of BF-PPO remain above those of CARDIFF for the entire training period. This shows that our method produces more effective defender strategies at every point in the training period, especially for the \texttt{MEANDER} attacker scenario.

We compare the converged strategies of BF-PPO with the published score of CARDIFF. The result is shown in Tab. \ref{tab:result}. BF-PPO achieves higher cumulative rewards for both attacker scenarios and for most time horizon, with significantly better performance against \texttt{MEANDER}. This observation is consistent with the performance gaps between the two methods in Fig. \ref{fig:learning-curves}.

The analysis of Fig. \ref{fig:learning-curves} and Tab. \ref{tab:result} leads us to the conclusion that BF-PPO provides an offline strategy learning method that performs better than CARDIFF, in both terms of convergence rate and convergence value.

%% file: parts/conclusion.tex
This paper formalises the CAGE-2 scenario using the POMDP framework. For this formal model, we define an optimal defender strategy and propose an iterative method to learn it. We call this method BF-PPO (Alg. \ref{alg:bf-ppo} and Alg. \ref{alg:bf-ppo-train}). It is based on the state-of-the-art reinforcement learning algorithm PPO. We use particle filter (Alg. \ref{alg:particle-filter}) to address the challenges of computational complexity due to the large state space of the CAGE-2 model. We evaluate BF-PPO for the CAGE-2 scenario and we find that our method outperforms CARDIFF, the highest ranked method on the CAGE-2 leaderboard. Our method produces a higher reward (Tab. \ref{tab:result}) and requires fewer training episodes than CARDIFF (Fig. \ref{fig:learning-curves}). 

The formal model developed in this paper allows for further investigation beyond computing approximate optimal strategies. For instance, an analysis of the state transitions of the model may allow us to find system configurations where an attacker cannot reach a certain target if the defender performs the correct actions. Second, given a fixed defender strategy, we can formulate an optimal attacker strategy and use our method BF-PPO to compute it. Also, we can formulate the interaction between attacker and defender as a game where both players follow dynamic strategies, and we can analyse and hopefully solve the game.

We are currently developing a different formalisation of CAGE-2, which uses causal modelling. The model captures the causal relations between key variables describing the network infrastructure as well as the attacker and the defender. It allows us to significantly reduce the policy search space of the corresponding solution method.

%% file: parts/appendix.tex

\section*{Appendix}
\label{sec:appendix}


\bgroup
\def\arraystretch{1.1} 
\begin{table}[htbp]
    \centering \footnotesize
    \caption{Configuration of the network infrastructure in CAGE-2.}    
    \begin{tabular}{p{0.09\textwidth}p{0.05\textwidth}p{0.07\textwidth}p{0.06\textwidth}p{0.08\textwidth}}
    \toprule
        \textbf{Host $h$} & \textbf{Subnet $z(h)$} & \textbf{Services} $E_h$ & \textbf{Access} $t(h,e)$ &  \textbf{Connectivity} $g_M(h)$  \\
    \midrule

    \multirow{2}{*}{CLIENT-1} & \multirow{2}{*}{1} & SSH & \textstate{S} & \multirow{2}{*}{ENT-1} \\
    & & FTP & \textstate{U} &  \\
    \midrule

    \multirow{2}{*}{CLIENT-2} & \multirow{2}{*}{1} & SMB & \textstate{N} & \multirow{2}{*}{ENT-1} \\
    & & RDS & \textstate{S} &  \\
    \midrule

    \multirow{3}{*}{CLIENT-3} & \multirow{3}{*}{1} & MYSQL & \textstate{S} & \multirow{3}{*}{ENT-0} \\
    & & APACHE2 & \textstate{U} &  \\
    & & SMTP & \textstate{S} &  \\
    \midrule

    \multirow{4}{*}{CLIENT-4} & \multirow{4}{*}{1} & SSHD & \textstate{S} & \multirow{4}{*}{ENT-0} \\
    & & MYSQL & \textstate{S} &  \\
    & & APACHE2 & \textstate{U} &  \\
    & & SMTP & \textstate{S} &  \\
    \midrule

    ENT-0 & 2 & SSHD & \textstate{S} & - \\
    \midrule

    \multirow{4}{*}{ENT-1} & \multirow{4}{*}{2} & SSHD & \textstate{S} & \multirow{4}{*}{-} \\
    & & RDS & \textstate{N} &  \\
    & & SMB & \textstate{S} &  \\
    & & TOMCAT8 & \textstate{U} &  \\
    \midrule

    \multirow{4}{*}{ENT-2} & \multirow{4}{*}{2} & SSHD & \textstate{S} & \multirow{4}{*}{OP-SERVER} \\
    & & RDS & \textstate{N} &  \\
    & & SMB & \textstate{S} &  \\
    & & TOMCAT8 & \textstate{U} &  \\
    \midrule

    OP-SERVER & 3 & SSHD & \textstate{S} & - \\
    \midrule

    OP-HOST-0 & 3 & SSHD & \textstate{S} & - \\
    \midrule

    OP-HOST-1 & 3 & SSHD & \textstate{S} & - \\
    \midrule

    OP-HOST-2 & 3 & SSHD & \textstate{S} & - \\
    \bottomrule
    
    \end{tabular}
    \label{tab:appendix-infrastructure} 
\end{table}
\egroup


\bgroup
\def\arraystretch{1.5} 
\begin{table}[ht]
    \centering
    \caption{Reward parameters for the defender in CAGE-2 in equation (\ref{eq:pomdp:reward})}
    \begin{tabular}{ccp{0.25\textwidth}}
    \toprule
        \textbf{Parameter} & \textbf{Value} & \textbf{Description}  \\
    \midrule 
    $\sigma_{\textaction{D_t=R}}$ & -1 & cost for performing action $D_t = \textaction{R}$ \\
    \midrule
    $\sigma_{\textaction{D_t \ne R}}$ & 0 & cost for performing other actions \\
    \midrule

    $\sigma_{z=1}$ & -0.1 & cost for each host in one of the states $\{\textstate{U}, \textstate{R}, \textstate{P}, \textstate{I}\}$ in the client subnet $z=1$ \\
    \midrule
    $\sigma_{z=2}$ & -1 & cost for each host in one of the states $\{\textstate{U}, \textstate{R}, \textstate{P}, \textstate{I}\}$ in the ENT subnet $z=2$ \\
    \midrule
    $\sigma_{z=3}$ & -1 & cost for each host in one of the states $\{\textstate{U}, \textstate{R}, \textstate{P}, \textstate{I}\}$ in operational subnet $z=3$ \\
    \midrule
    
    $\sigma_{h = \text{OP-SERVER}}$ & -10 & cost for interrupted services on OP-SERVER \\
    \midrule
    $\sigma_{h \neq \text{OP-SERVER}}$ & 0 & cost for interrupted services on other host \\
    \bottomrule
    \end{tabular}
    \label{tab:appendix-reward-param} 
\end{table}
\egroup

\bgroup
\def\arraystretch{1.5} 
\begin{table}[ht]
    \centering
    \caption{Hyperparameters for the training of defender strategies}
    \begin{tabular}{p{0.22\textwidth}p{0.2\textwidth}}
    \toprule
        \textbf{Parameter} & \textbf{Value} \\ 
    \midrule

    \multicolumn{2}{l}{Training configuration} \\
    \midrule
    Random seeds & $[0, 108, 153, 701]$ \\
    Time horizon $T$ & 100 \\
    \# iterations, \# episodes/iteration & 400, 100 \\ 
    \midrule

    \multicolumn{2}{l}{Neural network policy and PPO parameters} \\
    \midrule
    \# hidden layer, \# neurons/layer & 2, 64 \\
    Learning rate & $5 \cdot 10^{-4}$ \\
    \# epochs, discount factor  & $10, 0.99$ \\
    Optimiser (parameters) & Adam ($\beta_1 = 0.9, \beta_2 = 0.99)$ \\
    Clipping parameter $\epsilon$ & $0.2$ \\
    
    \bottomrule

    \end{tabular}
    \label{tab:appendix-hyperparam} 
\end{table}
\egroup

%% file: refs.bib
@inproceedings{snortids,
  author    = {Martin Roesch},
  title     = {Snort - Lightweight Intrusion Detection for Networks},
  booktitle = {Proceedings of the 13th USENIX Conference on System Administration (LISA)},
  year      = {1999},
  pages     = {229--238},
  publisher = {USENIX Association},
  address   = {Seattle, WA, USA}
}

@article{broids,
title = {Bro: a system for detecting network intruders in real-time},
journal = {Computer Networks},
volume = {31},
number = {23},
pages = {2435-2463},
year = {1999},
issn = {1389-1286},
doi = {https://doi.org/10.1016/S1389-1286(99)00112-7},
author = {Vern Paxson},
}

@misc{cyborg,
      title={CybORG: A Gym for the Development of Autonomous Cyber Agents}, 
      author={Maxwell Standen and Martin Lucas and David Bowman and Toby J. Richer and Junae Kim and Damian Marriott},
      year={2021},
      eprint={2108.09118},
      archivePrefix={arXiv},
      primaryClass={cs.CR},
}

@book{particle_filter,
  title={Probabilistic Robotics},
  author={Thrun, S. and Burgard, W. and Fox, D.},
  isbn={9780262201629},
  lccn={2005043346},
  year={2005},
  publisher={MIT Press}
}

@misc{ppo,
      title={Proximal Policy Optimization Algorithms}, 
      author={John Schulman and Filip Wolski and Prafulla Dhariwal and Alec Radford and Oleg Klimov},
      year={2017},
      eprint={1707.06347},
      archivePrefix={arXiv},
      primaryClass={cs.LG},
}

@article{cage2,
  title={On Autonomous Agents in a Cyber Defence Environment},
  author={Kiely, Mitchell and Bowman, David and Standen, Maxwell and Moir, Christopher},
  journal={arXiv preprint arXiv:2309.07388},
  year={2023}
}

@misc{kim-cage2,
      title={Optimal Defender Strategies for CAGE-2 using Causal Modeling and Tree Search}, 
      author={Kim Hammar and Neil Dhir and Rolf Stadler},
      year={2024},
      eprint={2407.11070},
      archivePrefix={arXiv},
      primaryClass={cs.LG},
}

@ARTICLE{gametheory1,
  author={Lau, Pikkin and Wei, Wei and Wang, Lingfeng and Liu, Zhaoxi and Ten, Chee-Wooi},
  journal={IEEE Transactions on Smart Grid}, 
  title={A Cybersecurity Insurance Model for Power System Reliability Considering Optimal Defense Resource Allocation}, 
  year={2020},
  volume={11},
  number={5},
  pages={4403-4414},
  doi={10.1109/TSG.2020.2992782}}

@ARTICLE{gametheory2,
  author={Hammar, Kim and Stadler, Rolf},
  journal={IEEE Transactions on Network and Service Management}, 
  title={Learning Near-Optimal Intrusion Responses Against Dynamic Attackers}, 
  year={2024},
  volume={21},
  number={1},
  pages={1158-1177},
  doi={10.1109/TNSM.2023.3293413}}

@ARTICLE{gametheory3,
  author={Zhang, Lefeng and Zhu, Tianqing and Hussain, Farookh Khadeer and Ye, Dayong and Zhou, Wanlei},
  journal={IEEE Transactions on Information Forensics and Security}, 
  title={A Game-Theoretic Method for Defending Against Advanced Persistent Threats in Cyber Systems}, 
  year={2023},
  volume={18},
  number={},
  pages={1349-1364},
  doi={10.1109/TIFS.2022.3229595}}

@INPROCEEDINGS{rl2,
  author={Hammar, Kim and Stadler, Rolf},
  booktitle={2021 17th International Conference on Network and Service Management (CNSM)}, 
  title={Learning Intrusion Prevention Policies through Optimal Stopping}, 
  year={2021},
  pages={509-517},
  doi={10.23919/CNSM52442.2021.9615542}}

@article{rl3,
author = {Tang, Yunlong and Sun, Jing and Wang, Huan and Deng, Junyi and Tong, Liang and Xu, Wenhong},
title = {A method of network attack-defense game and collaborative defense decision-making based on hierarchical multi-agent reinforcement learning},
year = {2024},
issue_date = {Jul 2024},
publisher = {Elsevier Advanced Technology Publications},
address = {GBR},
volume = {142},
number = {C},
issn = {0167-4048},
doi = {10.1016/j.cose.2024.103871},
journal = {Comput. Secur.},
month = jul,
numpages = {15},
}

@misc{causal1,
      title={Developing Optimal Causal Cyber-Defence Agents via Cyber Security Simulation}, 
      author={Alex Andrew and Sam Spillard and Joshua Collyer and Neil Dhir},
      year={2022},
      eprint={2207.12355},
      archivePrefix={arXiv},
      primaryClass={cs.CR},
}

@ARTICLE{causal2,
  author={Shi, Dawei and Guo, Ziyang and Johansson, Karl Henrik and Shi, Ling},
  journal={IEEE Transactions on Automatic Control}, 
  title={Causality Countermeasures for Anomaly Detection in Cyber-Physical Systems}, 
  year={2018},
  volume={63},
  number={2},
  pages={386-401},
  doi={10.1109/TAC.2017.2714646}}

@misc{related-work-cage-1,
      title={Beyond CAGE: Investigating Generalization of Learned Autonomous Network Defense Policies}, 
      author={Melody Wolk and Andy Applebaum and Camron Dennler and Patrick Dwyer and Marina Moskowitz and Harold Nguyen and Nicole Nichols and Nicole Park and Paul Rachwalski and Frank Rau and Adrian Webster},
      year={2022},
      eprint={2211.15557},
      archivePrefix={arXiv},
      primaryClass={cs.LG},
}

@inproceedings{related-work-cage-2,
   title={Autonomous Network Defence using Reinforcement Learning},
   DOI={10.1145/3488932.3527286},
   booktitle={Proceedings of the 2022 ACM on Asia Conference on Computer and Communications Security},
   publisher={ACM},
   author={Foley, Myles and Hicks, Chris and Highnam, Kate and Mavroudis, Vasilios},
   year={2022},
   month=may, collection={ASIA CCS ’22} }

@inproceedings{related-work-cage-3,
  author = {Foley, Myles and Wang, Mia and M, Zoe and Hicks, Chris and Mavroudis, Vasilios},
  booktitle = {CAMLIS},
  editor = {Raff, Edward and Samtani, Sagar and Deason, Lauren},
  ee = {https://ceur-ws.org/Vol-3391/paper1.pdf},
  pages = {1-19},
  publisher = {CEUR-WS.org},
  series = {CEUR Workshop Proceedings},
  title = {Inroads into Autonomous Network Defence using Explained Reinforcement Learning.},
  volume = 3391,
  year = 2022
}

@inproceedings{related-work-cage-4,
author = {Heckel, Kade},
title = {Neuroevolution for Autonomous Cyber Defense},
year = {2023},
isbn = {9798400701207},
publisher = {Association for Computing Machinery},
address = {New York, NY, USA},
doi = {10.1145/3583133.3590596},
booktitle = {Proceedings of the Companion Conference on Genetic and Evolutionary Computation},
pages = {651–654},
numpages = {4},
location = {Lisbon, Portugal},
series = {GECCO '23 Companion}
}

@inproceedings{related-work-cage-7,
  title={Cognitive models of dynamic decision in autonomous intelligent cyber defense},
  author={Prebot, Baptiste and Du, Yinuo and Xi, Xiaoli and Gonzalez, Cleotilde},
  booktitle={International Conference on Autonomous Intelligent Cyber-defense Agents},
  year={2022}
}

@INPROCEEDINGS{related-work-cage-jakob,
  author={Nyberg, Jakob and Johnson, Pontus and Méhes, András},
  booktitle={2022 IEEE/IFIP Network Operations and Management Symposium (NOMS)}, 
  title={Cyber threat response using reinforcement learning in graph-based attack simulations}, 
  year={2022},
  volume={},
  number={},
  pages={1-4},
  doi={10.1109/NOMS54207.2022.9789835}}

@ARTICLE{related-work-ids-1,
  author={Kurt, Mehmet Necip and Ogundijo, Oyetunji and Li, Chong and Wang, Xiaodong},
  journal={IEEE Transactions on Smart Grid}, 
  title={Online Cyber-Attack Detection in Smart Grid: A Reinforcement Learning Approach}, 
  year={2019},
  volume={10},
  number={5},
  pages={5174-5185},
  doi={10.1109/TSG.2018.2878570}}

@article{related-work-intrusion-response-1,
   title={Adaptive Security Response Strategies Through Conjectural Online Learning},
   volume={20},
   ISSN={1556-6021},
   DOI={10.1109/tifs.2025.3558600},
   journal={IEEE Transactions on Information Forensics and Security},
   publisher={Institute of Electrical and Electronics Engineers (IEEE)},
   author={Hammar, Kim and Li, Tao and Stadler, Rolf and Zhu, Quanyan},
   year={2025},
   pages={4055–4070} }

@ARTICLE{related-work-intrusion-response-2,
  author={Miehling, Erik and Rasouli, Mohammad and Teneketzis, Demosthenis},
  journal={IEEE Transactions on Information Forensics and Security}, 
  title={A POMDP Approach to the Dynamic Defense of Large-Scale Cyber Networks}, 
  year={2018},
  volume={13},
  number={10},
  pages={2490-2505},
  doi={10.1109/TIFS.2018.2819967}}

@inproceedings{related-work-intrusion-response-3,
author = {Mcabee, Ashley and Tummala, Murali and Mceachen, John},
year = {2021},
month = {01},
pages = {},
title = {The use of partially observable Markov decision processes to optimally implement moving target defense},
doi = {10.24251/HICSS.2021.840}
}

@article{related-work-intrusion-response-4,
author = {Hu, Zhisheng and Zhu, Minghui and Liu, Peng},
title = {Adaptive Cyber Defense Against Multi-Stage Attacks Using Learning-Based POMDP},
year = {2020},
issue_date = {February 2021},
publisher = {Association for Computing Machinery},
address = {New York, NY, USA},
volume = {24},
number = {1},
issn = {2471-2566},
doi = {10.1145/3418897},
month = nov,
articleno = {6},
numpages = {25}
}

@InProceedings{related-work-intrusion-response-5,
author="Pisal, Kshitij
and Roychowdhury, Sayak",
editor="Giri, Debasis
and Mandal, Jyotsna Kumar
and Sakurai, Kouichi
and De, Debashis",
title="Cyber-Defense Mechanism Considering Incomplete Information Using POMDP",
booktitle="Proceedings of International Conference on Network Security and Blockchain Technology",
year="2022",
publisher="Springer Nature Singapore",
address="Singapore",
pages="3--17",
}

@book{pomdp, 
place={Cambridge}, 
title={Partially Observed Markov Decision Processes: From Filtering to Controlled Sensing}, 
publisher={Cambridge University Press}, 
author={Krishnamurthy, Vikram}, 
year={2016}}

@article{pomdp-complexity1,
 ISSN = {0364765X, 15265471},
 author = {Christos H. Papadimitriou and John N. Tsitsiklis},
 journal = {Mathematics of Operations Research},
 number = {3},
 pages = {441--450},
 publisher = {INFORMS},
 title = {The Complexity of Markov Decision Processes},
 urldate = {2025-03-10},
 volume = {12},
 year = {1987}
}

@article{pomdp-complexity2,
title = {On the complexity of partially observed Markov decision processes},
journal = {Theoretical Computer Science},
volume = {157},
number = {2},
pages = {161-183},
year = {1996},
issn = {0304-3975},
doi = {https://doi.org/10.1016/0304-3975(95)00158-1},
author = {Dima Burago and Michel {de Rougemont} and Anatol Slissenko},
}

@article{dynamic-programming-pomdp,
 ISSN = {0030364X, 15265463},
 author = {Edward J. Sondik},
 journal = {Operations Research},
 number = {2},
 pages = {282--304},
 publisher = {INFORMS},
 title = {The Optimal Control of Partially Observable Markov Processes Over the Infinite Horizon: Discounted Costs},
 volume = {26},
 year = {1978}
}

@inproceedings{cage-github,
  title={Ttcp cage challenge 2},
  author={CAGE},
  booktitle={AAAI-22 Workshop on Artificial Intelligence for Cyber Security (AICS)},
  year={2022},
  url={https://github.com/cage-challenge/cage-challenge-2}
}

@misc{cardiff,
      title={Automated Cyber Defence: A Review}, 
      author={Sanyam Vyas and John Hannay and Andrew Bolton and Professor Pete Burnap},
      year={2023},
      primaryClass={cs.CR},
        howpublished = {code \url{https://github.com/john-cardiff/-cyborg-cage-2}},
}
